\documentclass[
]{ceurart}

\usepackage[frozencache,cachedir=.]{minted}
\setminted{breaklines=true}

\begin{document}

\copyrightyear{2021}
\copyrightclause{Copyright for this paper by its authors.
  Use permitted under Creative Commons License Attribution 4.0
  International (CC BY 4.0).}

\conference{Forum for Information Retrieval Evaluation, December 13-17, 2021, India}

\title{IIITDWD-ShankarB@ Dravidian-CodeMixi-HASOC2021: mBERT based model for identification of offensive content in south Indian languages}

\author[1]{Shankar Biradar}[%
email=shankar@iiitdwd.ac.in,
]
\address[1]{Indian Institute of Information Technology Dharwad,
}

\author[2]{Sunil Saumya}[%
email=sunil.saumya@iiitdwd.ac.in,
]
\address[2]{Indian Institute of Information Technology Dharwad}

\author[3]{Arun Chauhan}[%
email=aruntakhur@gmail.com
]
\address[3]{Graphic Era University Dehradun}

\begin{abstract}
 In recent years, there has been a lot of focus on offensive content. The amount of offensive content generated by social media is increasing at an alarming rate. This created a greater need to address this issue than ever before. To address these issues, the organizers of “Dravidian-Code Mixed HASOC-2020” have created two challenges. Task 1 involves identifying offensive content in Malayalam data, whereas Task 2 includes Malayalam and Tamil Code Mixed Sentences. Our team participated in Task 2. In our suggested model, we experiment with multilingual BERT to extract features, and three different classifiers are used on extracted features. Our model received a weighted F1 score of 0.70 for Malayalam data and was ranked fifth; we also received a weighted F1 score of 0.573 for Tamil Code Mixed data and were ranked eleventh.
\end{abstract}


\maketitle

\section{Introduction}
The availability of smart phones and the internet has created a great deal of interest in social media among today's youth. These applications give a huge platform for users to connect with the outside world and share their ideas and opinions with others. With these benefits comes a disadvantage: under the name of freedom to express one on social media, many people misuse the platform. People tend to publish inflammatory content on social media. This inflammatory information typically targets a single person, a group of people, a particular faith, or a community \cite{chowdhury2020multi}. People generate objectionable content and aggressively propagate it on social media in the lack of regulatory restraints.  This type of material is produced for a variety of reasons, including commercial and political gain \cite{santos2018fighting}. If not identified, this type of content has the ability to disrupt social harmony and cause riots in society. Also, it has the potential to have a detrimental psychological influence on the readers. It can have a negative impact on people's emotions and conduct.  Therefore, identifying this type of content is critical; as a result, researchers, policymakers, and investors (stakeholders) are attempting to develop a dependable technique to identify offensive content on social media \cite{kumar2020proceedings}.

Various studies on hate speech, harmful content, and abusive language identification in social media have been conducted during the previous decade. The majority of these studies were focused on monolingual English content and a large amount of English language cuprous has been created \cite{mubarak2017abusive}.  But, people in countries with a complex culture and history, such as India, frequently use regional languages to generate inappropriate social media posts. Users typically mix their regional languages with English while creating such content. This type of text is known as code mixed text on social media. Hence we require efficient method to classify offensive content in Code-Mixed Indian languages. In this context, the “Dravidian-CodeMixed HASOC-2020” shared task provider has organized two tasks for detecting hate speech in Dravidian languages such as Malayalam and Tamil code-mixed data. Our team took part in Task No. 2, and this paper presents the working notes for our suggested model.

 The rest of the article is arranged in the following manner: Section 2 provides a brief summary of previous work, while Section 3 describes the proposed model in full. Section 4 concludes by providing information on the outcome.

\section{Literature review}
A large number of researchers and practitioners from industry and academia have been attracted to the subject of automatic identification of hostile and harmful speech. \cite{fortuna2018survey} Provides a high-level review of current state-of-the-art in offensive language identification and related issues, such as hate speech recognition. \cite{davidson2017automated} Developed a publicly accessible dataset for identifying offensive language in tweets by categorizing them as hate speech, offensive but not hate speech, or neither. Various machine learning models, such as Support Vector Machine (SVM) and logistic regression, were created utilizing various data properties, such as n-grams, TF-IDF, readability, and so on. \cite{al2020proceedings} Built a model with deep neural networks in combination with SVM for the detection of offensive content with the accomplishment of F1 score of 90. \cite{chakravarthi2020sentiment} Provided a code mixed data set for Malayalam-English language and attained a base line result of 75\% F1 score using BERT's transformer model.

Offensive content detection from tweets is part of some conferences as challenging tasks. Offensive 2020, a task in five languages, was provided by SemEval in 2020 as a task in five languages, namely English, Arabic, Danish, Greek, and Turkish \cite{zampieri2020semeval}. In FIRE 2019, a similar task was achieved for Indo-European languages such as English, Hindi, and German. The data set was created using samples obtained on Twitter and Face book in all three languages. Various models, including LSTM with attention, Word2vec embedding with CNN, and BERT, were used for this task. In several cases, traditional learning models, outperformed deep learning methods for language other than English \cite{mandl2019overview}.

\section{Task and Data set description}
We have taken data set from HASOC subtask, offensive language identification of Dravidian CodeMix\cite{HASOC-dravidiancodemix-2021}. Challenges provided by the organizers are as follows.

Task 1: It is a binary classification problem with message level labeling for offensive and non-offensive information in Malayalam CodeMixed YouTube comments. 

Task 2: Given Romanized Tanglish and Manglish tweeter or YouTube comments, the system must classify them as offensive or non-offensive.

Our team took part in Task 2 for identifying offensive information in the Tanglish and Manglish data sets. According to organizer Tanglish data is collected from twitter tweets and comment on hello APP. Whereas Manglish data is taken from YouTube comments \cite{HASOC-dravidiancodemix-2021}. Detailed description of the data set is provided in Table \ref{data}, both Tanglish and Manglish data contain ID, Tweet and Label fields.
% \usepackage{multirow}
\begin{table}[t]
\caption{data set description}
\label{data}
\begin{tabular}{|l|l|l|l|l|}
\hline
\textbf{Task}                      & \textbf{Data set} & \textbf{Offensive} & \textbf{Not-offensive} & \textbf{Total} \\ \hline
\multirow{2}{*}{\textbf{Tanglish}} & Train             & 1980               & 2020                   & 4000           \\ \cline{2-5} 
                                   & Test              & 475                & 465                    & 940            \\ \hline
\multirow{2}{*}{\textbf{Manglish}} & Train             & 1953               & 2047                   & 4000           \\ \cline{2-5} 
                                   & Test              & 512                & 488                    & 1000           \\ \hline
\end{tabular}

\end{table}

\section{Methdology}
Our team has proposed three submissions based on three different models, each of which is designed using the general architecture shown in fig 1. Our model consists of three stages, each of which is discussed in the preceding subsections.
\begin{figure}[h!]
    \centering
    \includegraphics[scale=0.5]{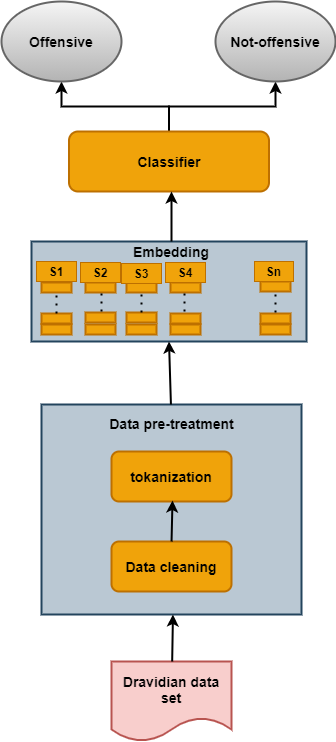}
    \caption{General architecture of classifier model}
    \label{fig1}
\end{figure}

\subsection{Data processing} data set provided by the organizer contains many unwanted information, few data preprocessing steps were undertaken on both text and label fields to convert the data suitable for model building. Digits, special characters, hyperlinks, and tweeter user handles were omitted from the data set because they were not helping us improve the performance of our model. Furthermore, the social media data provided by the organizer did not follow grammatical norms; hence, data lemmatization is performed to convert the data to its useable base form. Example, word ate, eaten and eating are converted to its base form eat. Converting text to lower case is also done. All of this preprocessing was done with the help of the NLTK toolbox from the Python library \cite{bird2009natural}. The preprocessed data is then fed into a tokenizer, which divides the tweet into a number of tokens. The mBERT tokanizer \footnote{https://huggingface.co/bert-base-multilingual-cased } is used for this purpose. Padding and masking were also used to handle variable length sentences.

\subsection{Feature extraction}
To obtain contextual embeddings from CodeMixed data, we employed the multilingual Bidirectional Encoder Representation (mBERT) model \cite{pires2019multilingual}. The architecture of the mBERT model is largely based on the original BERT architecture \cite{devlin2018bert}, which has 12 transformer blocks, 12 attention heads, and 768 hidden layers. Furthermore, the vector dimension of mBERT embeddings is 768. This model was trained using the same pre-training technique as the BERT, namely Masked Language Modeling (MLM) and Next Sentence Prediction. The only distinction is that multilingual BERT is trained on Wikipedia data from 104 different languages in order to handle languages other than English.  For our classification purposes, we only draw embeddings from the CLS token at the beginning because it gives whole sentence embeddings.

\subsection{Classification}
In our suggested model we experiment with three different classifiers the description of these models is provided in following subsections

\subsubsection{Traditional machine learning based classifier}
The initial studies were carried out utilizing typical machine algorithms such as Support Vector Machine (SVM) with a 10 fold cross validation. The experimental findings for the proposed model show that hyper-parameters of kernel value '1' and solver 'lbfgs' yield best results. This model takes an input vector from mBERT and produces labels that are either offensive or non-offensive. Python's scikit-learn library is used to implement the model \cite{pedregosa2011scikit}.

\subsubsection{Deep neural network based model}
Later, we experimented with the DNN, which in our suggested strategy serves as a second model. The DNN model consists of several dense layers designed to significantly shape and compress the input. Each dense layer follows a drop-out layer to prevent over fitting problem. We also have a batch normalization layer to normalize activation values. Embeddings obtained from mBERT is given as input this classifier.

\subsubsection{Transformer model}
We explored on transformer-based models in our last model. Transformer architectures are trained on generic tasks such as modeling language and can then be perfectly adjusted for classification purposes. BERT developers provide simple classification API, I’e Bert-base-uncased is underlying model for our classification.

\section{Experimental Results}

\begin{table}[t!]
\caption{Top performing models on Manglish data set}
\label{table2}
\begin{tabular}{|l|l|l|l|l|}
\hline
\textbf{Team name}      & \textbf{Precision} & \textbf{Recall} & \textbf{F1 score} & \textbf{Rank} \\ \hline
AIML                    & 0.776              & 0.762           & 0.766             & 1             \\ \hline
MUCIC                   & 0.764              & 0.76            & 0.762             & 2             \\ \hline
HSU                     & 0.744              & 0.73            & 0.735             & 3             \\ \hline
IIIT Surat              & 0.752              & 0.727           & 0.734             & 4             \\ \hline
IRLab                   & 0.754              & 0.705           & 0.714             & 5             \\ \hline
\textbf{IIITD-ShankarB} & \textbf{0.715}     & \textbf{0.693}  & \textbf{0.7}      & \textbf{6}    \\ \hline
\end{tabular}
\end{table}

To evaluate presented models, the organizers have provided a weighted F1 score. Among the proposed models, our top performing model came in sixth place for offensive content recognition in the Mangalish data set and eleventh place in the Tanglish data set, Table \ref{table2} and Table \ref{table3} provides the list of top performing models with weighted F1 score for Manglish and Tanglish data set respectively [The result of our proposed model is shown in bold letters]. We trained our proposed model on Tanglish data set of 4000 comments from the training set and tested on 940 comments from the test set. For the Manglish data set, 4000 train comments and 1000 test comments are employed.

\begin{table}[t!]
\caption{Top performing models on Tanglish data set}
\label{table3}
\begin{tabular}{|l|l|l|l|l|}
\hline
\textbf{Team name}      & \textbf{Precision} & \textbf{Recall} & \textbf{F1 score} & \textbf{Rank} \\ \hline
MUCIC                   & 0.679              & 0.685           & 0.678             & 1             \\ \hline
AIML 2                  & 0.67               & 0.67            & 0.67              & 2             \\ \hline
SSN\_IT\_NLP            & 0.685              & 0.688           & 0.668             & 3             \\ \hline
ZYBank AI               & 0.671              & 0.676           & 0.654             & 4             \\ \hline
IRLab                   & 0.654              & 0.662           & 0.65              & 5             \\ \hline
\textbf{IIITD-ShankarB} & \textbf{0.599}     & \textbf{0.568}  & \textbf{0.573}    & \textbf{11}   \\ \hline
\end{tabular}
\end{table}

\section{Conclusion and future enhancement}
In our article we presented model submitted by our team IIITD-ShankarB for offensive content identification in shared task “Dravidian-CodeMixed HASOC-2020”. In our proposed work, we experimented with three distinct models, including a machine learning-based model, a deep Neural Network model, and a transformer-based language model. Our model is one of the top performing models, ranking sixth on the Manglish data set and eleventh on the Tanglish data set. In future work, we can improve the efficiency of the suggested model by including domain-specific embeddings.

%%
%% Define the bibliography file to be used
\bibliography{sample.bib}

\begin{thebibliography}{15}
\expandafter\ifx\csname natexlab\endcsname\relax\def\natexlab#1{#1}\fi
\providecommand{\url}[1]{\texttt{#1}}
\providecommand{\href}[2]{#2}
\providecommand{\path}[1]{#1}
\providecommand{\DOIprefix}{doi:}
\providecommand{\ArXivprefix}{arXiv:}
\providecommand{\URLprefix}{URL: }
\providecommand{\Pubmedprefix}{pmid:}
\providecommand{\doi}[1]{\href{http://dx.doi.org/#1}{\path{#1}}}
\providecommand{\Pubmed}[1]{\href{pmid:#1}{\path{#1}}}
\providecommand{\bibinfo}[2]{#2}
\ifx\xfnm\relax \def\xfnm[#1]{\unskip,\space#1}\fi
%Type = Inproceedings
\bibitem[{Chowdhury et~al.(2020)Chowdhury, Mubarak, Abdelali, Jung, Jansen, and
  Salminen}]{chowdhury2020multi}
\bibinfo{author}{S.~A. Chowdhury}, \bibinfo{author}{H.~Mubarak},
  \bibinfo{author}{A.~Abdelali}, \bibinfo{author}{S.-g. Jung},
  \bibinfo{author}{B.~J. Jansen}, \bibinfo{author}{J.~Salminen},
\newblock \bibinfo{title}{A multi-platform arabic news comment dataset for
  offensive language detection},
\newblock in: \bibinfo{booktitle}{Proceedings of the 12th Language Resources
  and Evaluation Conference}, \bibinfo{year}{2020}, pp.
  \bibinfo{pages}{6203--6212}.
%Type = Article
\bibitem[{Santos et~al.(2018)Santos, Melnyk, and Padhi}]{santos2018fighting}
\bibinfo{author}{C.~N.~d. Santos}, \bibinfo{author}{I.~Melnyk},
  \bibinfo{author}{I.~Padhi},
\newblock \bibinfo{title}{Fighting offensive language on social media with
  unsupervised text style transfer},
\newblock \bibinfo{journal}{arXiv preprint arXiv:1805.07685}
  (\bibinfo{year}{2018}).
%Type = Inproceedings
\bibitem[{Kumar et~al.(2020)Kumar, Ojha, Lahiri, Zampieri, Malmasi, Murdock,
  and Kadar}]{kumar2020proceedings}
\bibinfo{author}{R.~Kumar}, \bibinfo{author}{A.~K. Ojha},
  \bibinfo{author}{B.~Lahiri}, \bibinfo{author}{M.~Zampieri},
  \bibinfo{author}{S.~Malmasi}, \bibinfo{author}{V.~Murdock},
  \bibinfo{author}{D.~Kadar},
\newblock \bibinfo{title}{Proceedings of the second workshop on trolling,
  aggression and cyberbullying},
\newblock in: \bibinfo{booktitle}{Proceedings of the Second Workshop on
  Trolling, Aggression and Cyberbullying}, \bibinfo{year}{2020}.
%Type = Inproceedings
\bibitem[{Mubarak et~al.(2017)Mubarak, Darwish, and Magdy}]{mubarak2017abusive}
\bibinfo{author}{H.~Mubarak}, \bibinfo{author}{K.~Darwish},
  \bibinfo{author}{W.~Magdy},
\newblock \bibinfo{title}{Abusive language detection on arabic social media},
\newblock in: \bibinfo{booktitle}{Proceedings of the first workshop on abusive
  language online}, \bibinfo{year}{2017}, pp. \bibinfo{pages}{52--56}.
%Type = Article
\bibitem[{Fortuna and Nunes(2018)}]{fortuna2018survey}
\bibinfo{author}{P.~Fortuna}, \bibinfo{author}{S.~Nunes},
\newblock \bibinfo{title}{A survey on automatic detection of hate speech in
  text},
\newblock \bibinfo{journal}{ACM Computing Surveys (CSUR)} \bibinfo{volume}{51}
  (\bibinfo{year}{2018}) \bibinfo{pages}{1--30}.
%Type = Inproceedings
\bibitem[{Davidson et~al.(2017)Davidson, Warmsley, Macy, and
  Weber}]{davidson2017automated}
\bibinfo{author}{T.~Davidson}, \bibinfo{author}{D.~Warmsley},
  \bibinfo{author}{M.~Macy}, \bibinfo{author}{I.~Weber},
\newblock \bibinfo{title}{Automated hate speech detection and the problem of
  offensive language},
\newblock in: \bibinfo{booktitle}{Proceedings of the International AAAI
  Conference on Web and Social Media}, volume~\bibinfo{volume}{11},
  \bibinfo{year}{2017}.
%Type = Inproceedings
\bibitem[{Al-Khalifa et~al.(2020)Al-Khalifa, Magdy, Darwish, Elsayed, and
  Mubarak}]{al2020proceedings}
\bibinfo{author}{H.~Al-Khalifa}, \bibinfo{author}{W.~Magdy},
  \bibinfo{author}{K.~Darwish}, \bibinfo{author}{T.~Elsayed},
  \bibinfo{author}{H.~Mubarak},
\newblock \bibinfo{title}{Proceedings of the 4th workshop on open-source arabic
  corpora and processing tools, with a shared task on offensive language
  detection},
\newblock in: \bibinfo{booktitle}{Proceedings of the 4th Workshop on
  Open-Source Arabic Corpora and Processing Tools, with a Shared Task on
  Offensive Language Detection}, \bibinfo{year}{2020}.
%Type = Article
\bibitem[{Chakravarthi et~al.(2020)Chakravarthi, Jose, Suryawanshi, Sherly, and
  McCrae}]{chakravarthi2020sentiment}
\bibinfo{author}{B.~R. Chakravarthi}, \bibinfo{author}{N.~Jose},
  \bibinfo{author}{S.~Suryawanshi}, \bibinfo{author}{E.~Sherly},
  \bibinfo{author}{J.~P. McCrae},
\newblock \bibinfo{title}{A sentiment analysis dataset for code-mixed
  malayalam-english},
\newblock \bibinfo{journal}{arXiv preprint arXiv:2006.00210}
  (\bibinfo{year}{2020}).
%Type = Article
\bibitem[{Zampieri et~al.(2020)Zampieri, Nakov, Rosenthal, Atanasova,
  Karadzhov, Mubarak, Derczynski, Pitenis, and
  {\c{C}}{\"o}ltekin}]{zampieri2020semeval}
\bibinfo{author}{M.~Zampieri}, \bibinfo{author}{P.~Nakov},
  \bibinfo{author}{S.~Rosenthal}, \bibinfo{author}{P.~Atanasova},
  \bibinfo{author}{G.~Karadzhov}, \bibinfo{author}{H.~Mubarak},
  \bibinfo{author}{L.~Derczynski}, \bibinfo{author}{Z.~Pitenis},
  \bibinfo{author}{{\c{C}}.~{\c{C}}{\"o}ltekin},
\newblock \bibinfo{title}{Semeval-2020 task 12: Multilingual offensive language
  identification in social media (offenseval 2020)},
\newblock \bibinfo{journal}{arXiv preprint arXiv:2006.07235}
  (\bibinfo{year}{2020}).
%Type = Inproceedings
\bibitem[{Mandl et~al.(2019)Mandl, Modha, Majumder, Patel, Dave, Mandlia, and
  Patel}]{mandl2019overview}
\bibinfo{author}{T.~Mandl}, \bibinfo{author}{S.~Modha},
  \bibinfo{author}{P.~Majumder}, \bibinfo{author}{D.~Patel},
  \bibinfo{author}{M.~Dave}, \bibinfo{author}{C.~Mandlia},
  \bibinfo{author}{A.~Patel},
\newblock \bibinfo{title}{Overview of the hasoc track at fire 2019: Hate speech
  and offensive content identification in indo-european languages},
\newblock in: \bibinfo{booktitle}{Proceedings of the 11th forum for information
  retrieval evaluation}, \bibinfo{year}{2019}, pp. \bibinfo{pages}{14--17}.
%Type = Inproceedings
\bibitem[{Chakravarthi et~al.(2021)Chakravarthi, Kumaresan, Sakuntharaj,
  Madasamy, Thavareesan, B, Chinnaudayar~Navaneethakrishnan, McCrae, and
  Mandl}]{HASOC-dravidiancodemix-2021}
\bibinfo{author}{B.~R. Chakravarthi}, \bibinfo{author}{P.~K. Kumaresan},
  \bibinfo{author}{R.~Sakuntharaj}, \bibinfo{author}{A.~K. Madasamy},
  \bibinfo{author}{S.~Thavareesan}, \bibinfo{author}{P.~B},
  \bibinfo{author}{S.~Chinnaudayar~Navaneethakrishnan}, \bibinfo{author}{J.~P.
  McCrae}, \bibinfo{author}{T.~Mandl},
\newblock \bibinfo{title}{{Overview of the HASOC-DravidianCodeMix Shared Task
  on Offensive Language Detection in Tamil and Malayalam}},
\newblock in: \bibinfo{booktitle}{Working Notes of FIRE 2021 - Forum for
  Information Retrieval Evaluation}, \bibinfo{publisher}{CEUR},
  \bibinfo{year}{2021}.
%Type = Book
\bibitem[{Bird et~al.(2009)Bird, Klein, and Loper}]{bird2009natural}
\bibinfo{author}{S.~Bird}, \bibinfo{author}{E.~Klein},
  \bibinfo{author}{E.~Loper}, \bibinfo{title}{Natural language processing with
  Python: analyzing text with the natural language toolkit},
  \bibinfo{publisher}{" O'Reilly Media, Inc."}, \bibinfo{year}{2009}.
%Type = Article
\bibitem[{Pires et~al.(2019)Pires, Schlinger, and
  Garrette}]{pires2019multilingual}
\bibinfo{author}{T.~Pires}, \bibinfo{author}{E.~Schlinger},
  \bibinfo{author}{D.~Garrette},
\newblock \bibinfo{title}{How multilingual is multilingual bert?},
\newblock \bibinfo{journal}{arXiv preprint arXiv:1906.01502}
  (\bibinfo{year}{2019}).
%Type = Article
\bibitem[{Devlin et~al.(2018)Devlin, Chang, Lee, and
  Toutanova}]{devlin2018bert}
\bibinfo{author}{J.~Devlin}, \bibinfo{author}{M.-W. Chang},
  \bibinfo{author}{K.~Lee}, \bibinfo{author}{K.~Toutanova},
\newblock \bibinfo{title}{Bert: Pre-training of deep bidirectional transformers
  for language understanding},
\newblock \bibinfo{journal}{arXiv preprint arXiv:1810.04805}
  (\bibinfo{year}{2018}).
%Type = Article
\bibitem[{Pedregosa et~al.(2011)Pedregosa, Varoquaux, Gramfort, Michel,
  Thirion, Grisel, Blondel, Prettenhofer, Weiss, Dubourg
  et~al.}]{pedregosa2011scikit}
\bibinfo{author}{F.~Pedregosa}, \bibinfo{author}{G.~Varoquaux},
  \bibinfo{author}{A.~Gramfort}, \bibinfo{author}{V.~Michel},
  \bibinfo{author}{B.~Thirion}, \bibinfo{author}{O.~Grisel},
  \bibinfo{author}{M.~Blondel}, \bibinfo{author}{P.~Prettenhofer},
  \bibinfo{author}{R.~Weiss}, \bibinfo{author}{V.~Dubourg}, et~al.,
\newblock \bibinfo{title}{Scikit-learn: Machine learning in python. the journal
  of machine learning research 12}  (\bibinfo{year}{2011}).

\end{thebibliography}

\end{document}